\documentclass[11pt,a4paper]{article}
\usepackage{times,latexsym}
\usepackage{url}
\usepackage[T1]{fontenc}
\usepackage{amssymb,amsmath,amsthm,enumitem}
\usepackage{algorithm}
\usepackage[a-1b]{pdfx}
\usepackage[noend]{algpseudocode}

\usepackage{microtype}

\usepackage{tikz}
\usetikzlibrary{shapes.geometric}
\usetikzlibrary{patterns}
\usepackage{pgfplots}
\usetikzlibrary{backgrounds}
\usetikzlibrary{matrix,calc}
\usetikzlibrary{svg.path}
\usetikzlibrary{automata,positioning,decorations.text,topaths,arrows.meta,decorations.pathmorphing,quotes}
\usepackage{booktabs}
\usepackage{xcolor}

\pgfplotsset{nice plot/.style={
            every axis plot post/.style={/pgf/number format/fixed},
            width=0.48\textwidth,
            height=3cm,
            ymajorgrids=false,
            yminorgrids=false,
            xtick={0,1,2,3,4,5},
            xticklabels={0,1,2,3,4,5},
            every x tick label/.append style={font=\tiny},
            every y tick label/.append style={font=\tiny},
            ymin=0,
            tick pos=left,
            axis y line*=left,
            axis x line*=bottom,
            symbolic x coords={0,1,2,3,4,5},
            ylabel near ticks,
            ylabel shift={-3pt},
            xlabel near ticks,
            xlabel shift={-3pt},
            enlarge x limits=0.1,
            title style={yshift=-.1cm,font=\small},
    }
}
\newcommand{\niceplot}[5]{
\begin{tikzpicture}
    \begin{axis}[nice plot,
    title={#3},ylabel={#4},xlabel={#5}]
    \addplot [mark=none,nodes near coords, thick, densely dashed,
            every node near coord/.append style={font=\tiny,color=black},] table[x=Iterations,y=#1]{\itdata};
    \addplot [only marks,mark=square*,mark options={fill=white,draw=white},mark size=1pt,] table[x=Iterations,y=#1]{\itdata};
    \addplot [only marks,mark=o,mark options={fill=blue,draw=blue},mark size=.3pt,] table[x=Iterations,y=#1]{\itdata};
    \addplot [mark=none,color=red,nodes near coords,every node near coord/.append style={font=\tiny,yshift=-10pt,color=red}] table[x=Iterations,y=#2]{\itdata};
    \addplot [only marks,mark=square*,mark options={fill=white,draw=white},mark size=1pt,] table[x=Iterations,y=#2]{\itdata};
    \addplot [only marks,mark=o,mark options={fill=red,draw=red},mark size=.3pt,] table[x=Iterations,y=#2]{\itdata};
    \end{axis}
\end{tikzpicture}
}

\usepackage[acceptedWithA]{tacl2018v2}

\usepackage{xspace,mfirstuc,tabulary}

\newif\iftaclinstructions
\taclinstructionsfalse 
\iftaclinstructions
\renewcommand{\confidential}{}
\renewcommand{\anonsubtext}{(No author info supplied here, for consistency with
TACL-submission anonymization requirements)}
\newcommand{\instr}
\fi

\iftaclpubformat 

\else

\fi

\newif\ifspace
\spacefalse
\definecolor{burntred}{HTML}{E66101}
\definecolor{burntblue}{HTML}{00A2FF}

\DeclareMathOperator*{\argmax}{arg\,max}

\newcommand{\ba}{`}

\renewcommand{\arraystretch}{1.1}

\title{Lexically-Aware Semi-Supervised Learning for OCR Post-Correction}

\author{
 Template Author \\
 Template Affiliation/Address Line 1 \\
 Template Affiliation/Address Line 2 \\
 Template Affiliation/Address Line 2 \\
  {\sf template.email@sampledomain.com} \\
}

\author{
    Shruti Rijhwani\textsuperscript{$1$}, Daisy Rosenblum\textsuperscript{$2$}, Antonios Anastasopoulos\textsuperscript{$3$}, Graham Neubig\textsuperscript{$1$}  \\
    \textsuperscript{$1$}Language Technologies Institute, Carnegie Mellon University\\
    \textsuperscript{$2$}University of British Columbia\\
    \textsuperscript{$3$}Department of Computer Science, George Mason University\\
  \texttt{srijhwan@cs.cmu.edu, daisy.rosenblum@ubc.ca,}\\ \texttt{antonis@gmu.edu, gneubig@cs.cmu.edu}
}

\date{}

\begin{document}
\maketitle
\begin{abstract}
Much of the existing linguistic data in many languages of the world is locked away in non-digitized books and documents.
Optical character recognition (OCR) can be used to produce digitized text, and previous work has demonstrated the utility of neural post-correction methods that improve the results of general-purpose OCR systems on recognition of less-well-resourced languages.
However, these methods rely on manually curated post-correction data, which are relatively scarce compared to the non-annotated raw images that need to be digitized.
In this paper, we present a semi-supervised learning method that makes it possible to utilize these raw images to improve performance, specifically through the use of \emph{self-training}, a technique where a model is iteratively trained on its own outputs.
In addition, to enforce consistency in the recognized vocabulary, we introduce a \emph{lexically-aware decoding} method that augments the neural post-correction model with a count-based language model constructed from the recognized texts, implemented using weighted finite-state automata (WFSA) for efficient and effective decoding.
Results on four endangered languages demonstrate the utility of the proposed method, with relative error reductions of 15-29\%, where we find the combination of self-training and lexically-aware decoding essential for achieving consistent improvements.\footnote{Data and code are available at \url{https://shrutirij.github.io/ocr-el/}.}
\end{abstract}

\section{Introduction}
\label{sec:intro}

\begin{figure}[tb]
    \centering
    \small
    \begin{tabular}{@{}lc@{}}
        \raisebox{1.2em}{[Image]} & {\includegraphics[width=0.45\columnwidth]{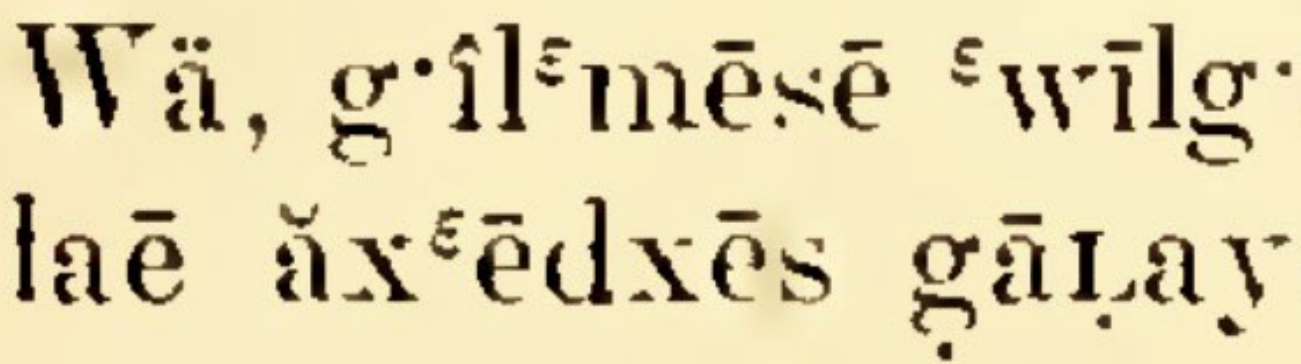}} \\
        & $\Big\downarrow$ \\
        \raisebox{1.2em}{[First pass OCR]} \hspace{0.75cm} & \raisebox{0.3em}{\includegraphics[width=0.45\columnwidth]{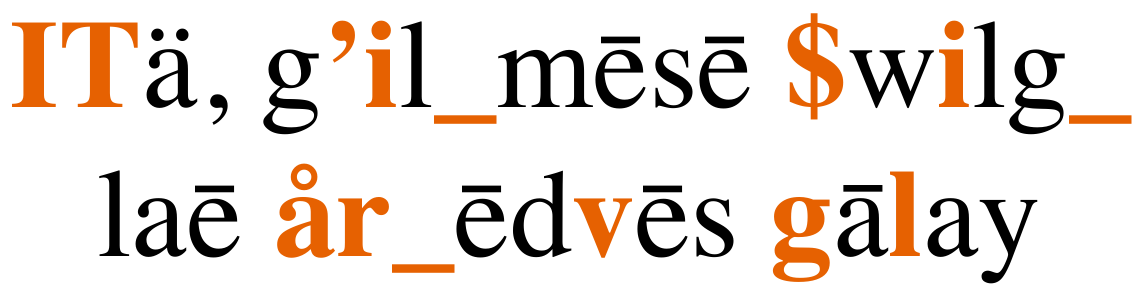}} \\
        & $\Big\downarrow$\\
        \raisebox{1.4em}{[Post-corrected]} & \raisebox{0.3em}{\includegraphics[width=0.40\columnwidth]{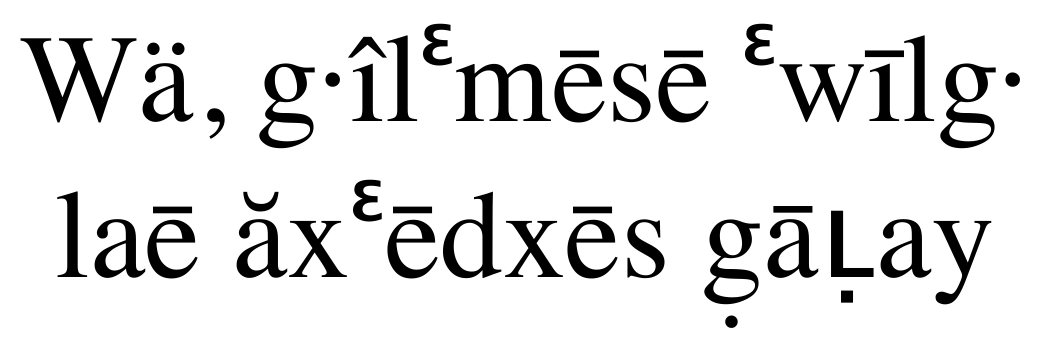}} \\
    \end{tabular}
    \caption{OCR post-correction on a scanned document that contains text in the endangered language Kwak'wala. The goal of post-correction is to fix the {\color{burntred} \textbf{recognition errors}} made by the first pass OCR system.}
    \label{fig:example}
    \ifspace
    \vspace{-1em}
    \fi
\end{figure}

There is a vast amount of textual data available in printed form~\cite{dong-smith-2018-multi}. In this paper, we address the task of digitizing printed materials that contain text in endangered languages, i.e., languages with small populations of first-language speakers and limited acquisition among younger speakers. Printed texts in endangered languages come from various sources, including linguistic documentation and cultural and educational books.

Extracting text data from these documents is valuable for a multitude of reasons. Automatic digitization can aid language documentation, preservation, and accessibility efforts by archiving the texts and making them searchable for language learners, teachers, and speakers, contributing to essential resources for community-based language revitalization. Further, most endangered languages are under-represented in natural language processing technologies, primarily because there is little to no data available for training and evaluation~\cite{joshi2020state}. This challenge can be mitigated by converting printed materials in these languages to a machine-readable format. 

Optical character recognition (OCR) systems can be used to produce digitized text, and recent work~\cite{rijhwani-etal-2020-ocr} has demonstrated that \textit{post-correction} improves the performance of existing general-purpose OCR systems on endangered languages (an example is in \autoref{fig:example}). Most state-of-the-art OCR post-correction methods use neural sequence-to-sequence models and rely on considerable resources such as a large number of manual transcriptions~\cite{schnober-etal-2016-still,icdar} or substantial textual data to train a language model~\cite{dong-smith-2018-multi}. To adapt these methods for the less-well-resourced endangered languages setting, \citet{rijhwani-etal-2020-ocr} add translations and structural biases to the model.

However, even with such methods targeted to low-resource learning, post-correction performance is still dependent on manually curated data, which are minimally available for most endangered languages. On the other hand, unannotated raw images that need to be digitized are relatively less scarce; for many endangered languages, hundreds of printed pages exist, with only a small subset manually transcribed. In this paper, we propose a semi-supervised learning method for OCR post-correction that efficiently utilizes these unannotated pages to improve performance.

The method has two key components. We first present a \textbf{self-training} method for OCR post-correction (\autoref{sec:selftraining}) to create \emph{pseudo-}training data. A baseline post-correction model is used to correct the initial OCR output on the unannotated pages, and the generated \ba\ba post-corrected'' text is then used as \emph{pseudo-}training data to improve the post-correction model. The self-training process is repeated to iteratively obtain better predictions on the unannotated pages.

While self-training is a straightforward way to use the unannotated data, incorrect predictions in the \emph{pseudo-}training data may introduce noise into the model~\cite{zhu2009introduction}. To counterbalance the influence of this noise, we propose \textbf{lexically-aware decoding} (\autoref{sec:lexical}), an inference strategy that encourages the model to generate predictions that contain \ba\ba known'' words. We use the \emph{pseudo-}training data to train a count-based language model, represented with a weighted finite-state automaton (WFSA). Our proposed decoding method jointly uses an LSTM decoder and the WFSA to make OCR post-correction predictions. 

The intuition behind the joint decoding strategy is simple. As the model iteratively improves with self-training, the quality of the \emph{pseudo-}training data is also likely to improve and contain an increasing number of correctly predicted words, resulting in a better count-based language model. Consequently, joint decoding \textit{reinforces} the prediction of more accurate words and mitigates the noise introduced by incorrect words in the \emph{pseudo-}training data.

We conduct experiments on four endangered languages: Ainu, Griko, Kwak'wala, and Yakkha. Our proposed method reduces the character and word error rates by 15\%--29\% over a state-of-the-art OCR post-correction method for endangered languages. We find that the \textit{combination} of self-training and lexically-aware decoding is essential for achieving consistent improvements in performance.

\section{Problem Formulation}
\label{sec:background}

\paragraph{Optical Character Recognition} The OCR task involves generating a transcription of the text contained in an image. In this paper, we use existing OCR tools (detailed in \autoref{sec:firstpass}) to obtain a \textit{first pass transcription} for the images in our dataset. The first pass transcription is a text sequence of $N$ characters, denoted as $\boldsymbol{x}=[x_1, \ldots, x_N]$.

\paragraph{OCR Post-Correction}
Even state-of-the-art OCR models are susceptible to making recognition errors~\cite{dong-smith-2018-multi}. 
Errors are particularly frequent in the case of endangered languages because most off-the-shelf OCR tools do not directly support these languages and training a high-performance OCR system is challenging given the small amount of data that is typically available~\cite{rijhwani-etal-2020-ocr}. We use OCR post-correction to correct these errors and improve the quality of the transcription.

The post-correction model takes the first pass transcription $\boldsymbol{x}$ as input and generates the \textit{corrected transcription}, a sequence of $T$ characters denoted as $\boldsymbol{y}=[y_1, \ldots, y_T]$:
$$\boldsymbol{y} = \argmax_{\boldsymbol{y'}} p_\text{corr}(\boldsymbol{y'}|\boldsymbol{x})$$

\section{Base Model}
\label{sec:base}

As the base post-correction model, we use the model from \citet{rijhwani-etal-2020-ocr}: a sequence-to-sequence model that uses an attention-based LSTM encoder-decoder~\cite{bahdanau2015neural}, with adaptations for low-resource OCR post-correction. We briefly describe the method here but refer readers to the original paper for details.

The OCR post-correction model takes the first pass transcription $\boldsymbol{x}$ as input, with the aim of predicting an error-free transcription $\boldsymbol{y}$. First, each character in the input sequence $\boldsymbol{x}$ is mapped to a vector representation using character embeddings. This forms a sequence of vectors, $\mathbf{x}=[\mathbf{x}_1, \ldots, \mathbf{x}_N]$. The \textbf{encoder} is a character-level bidirectional LSTM~\cite{hochreiter1997long}, which transforms $\mathbf{x}$ into a sequence of hidden state vectors $\mathbf{h}=[\mathbf{h}_1, \ldots, \mathbf{h}_N]$.\footnote{\citet{rijhwani-etal-2020-ocr} incorporate translations into the model with a multi-source encoder. We omit this from our formulation, considering applicability to texts without available translations. However, adding an encoder into our framework remains straightforward and can be used if translations exist.} 

The model's decoding process uses an \textbf{attention mechanism} to provide context from the encoder hidden states. At each decoding timestep $t$, the attention layer uses $\mathbf{h}$ and the decoder state from the previous timestep, $\mathbf{s}_{t-1}$, to produce the context vector $\mathbf{c}_t$. The LSTM \textbf{decoder}, given $\mathbf{c}_t$, computes the output state $\mathbf{s}_t$ and subsequently the probability distribution $\mathbf{y}_t$ for generating the next character of the target sequence $\boldsymbol{y}$:
\begin{equation}
p\left(\mathbf{y}_t\right) = \mathrm{softmax}\left(\mathbf{W}\mathbf{s}_t + \mathbf{b}\right)
\label{eq:decoder}
\end{equation}

\noindent
\citet{rijhwani-etal-2020-ocr} adapt the encoder-decoder model above for \textbf{low-resource post-correction} by adding pretraining and three structural biases: 
\begin{itemize}[leftmargin=*,itemsep=0pt]
    \item \textbf{Diagonal attention loss}: OCR post-correction is a monotonic sequence-to-sequence task. Hence, the attention weights are expected to be higher closer to the diagonal -- adding attention elements off the diagonal to the training loss encourages monotonic attention~\cite{cohn-etal-2016-incorporating}.
    \item \textbf{Copy mechanism}: The copy mechanism enables the model to choose between generating a character based on the decoder state (\autoref{eq:decoder}) or copying a character directly from the input sequence $\boldsymbol{x}$ by sampling from the attention distribution~\cite{gu-etal-2016-incorporating,see-etal-2017-get}.
    \item \textbf{Coverage}: The coverage vector keeps track of attention weights from previous timesteps. It is used as additional information when computing $\mathbf{c}_t$ and is added to the training loss to discourage the model from repeatedly attending to the same character~\cite{mi-etal-2016-coverage,tu-etal-2016-modeling}.
\end{itemize}

The model is trained in a \textbf{supervised} manner with a small number of manual transcriptions: the training data includes pairs of first pass OCR text with its corresponding error-free transcription. The \textbf{post-correction training loss function} (denoted as $\mathcal{L}$) is a combination of cross-entropy loss along with the diagonal attention loss and the coverage loss from the structural biases. Inference with a trained model is performed using \textbf{beam search}.

In the following sections, we use the method described above as a base model for our proposed semi-supervised learning technique for OCR post-correction. Given the minimal manually transcribed data we have in endangered languages, our approach aims to efficiently use the relatively larger number of pages without gold transcriptions to improve performance. To this end, we introduce two methodological improvements: (1) self-training and (2) lexically-aware decoding.

\section{Self-Training}
\label{sec:selftraining}
Self-training is a semi-supervised learning method, where a trained model is used to make predictions on unlabeled data, and the model is then retrained on its own predictions~\cite{zhu2009introduction}.

Consider that we have a set of images with manually created transcriptions and a set of images without gold transcriptions. We can obtain a first pass transcription for the text contained in the images (both sets) with existing OCR tools.

More formally, we have a gold-transcribed dataset $D=\{\langle\boldsymbol{x}^{(i)},\boldsymbol{y}^{(i)}\rangle\}_{i=1}^{d}$, where $\boldsymbol{x}^{(i)}$ is the first pass transcription and $\boldsymbol{y}^{(i)}$ is the error-free manual transcription of the $i$th training instance.\footnote{In our dataset, the source and target data instances are either at the line-level or the sentence-level (see \autoref{sec:dataset}).} We also have a dataset for which only the first pass OCR is available (i.e., no manual transcriptions), $U=\{\boldsymbol{x}^{(j)}\}_{j=1}^u$. For most cases in the endangered languages setting, the set without gold transcriptions is much larger, that is, $u \gg d$. 

Since self-training requires a baseline model to get an initial set of predictions on $U$, we first train the base model described in \autoref{sec:base} on the gold-transcribed set $D$. Let the trained base model be $f_\theta$. Next, we use the predictions on $U$ from $f_\theta$ to self-train the model. We follow the self-training strategy recommended in \citet{he2019revisiting}, which involves two steps: \ba\ba pseudo-training'' and \ba\ba fine-tuning''. We describe each step of the self-training procedure in detail below:

\begin{enumerate}[leftmargin=*,parsep=0.5em,itemsep=0pt]
    \item Apply the initial OCR post-correction model $f_\theta$ to each instance in the set $U$ to obtain predictions using beam search inference. 
    
    For an instance $\boldsymbol{x}$, let the prediction be $f_\theta(\boldsymbol{x})$.
    
    \item Create a \textit{pseudo-annotated dataset} with the predictions from step 1. Let this be $S = \{\langle\boldsymbol{x}, f_\theta(\boldsymbol{x})\rangle \mid \boldsymbol{x} \in U\}$.
    
    \item Train the model $f_\theta$ on sets $U$ and $S$.
    
    This is the \textit{pseudo-training} step. Here, we first train the encoder and the decoder components with a language modeling objective, and then train the end-to-end post-correction model. The procedure is as follows:
    
    \begin{enumerate}[label=\alph*),leftmargin=0.5em,parsep=0.5em]
        \item Train the encoder with a character-level language modeling (LM) objective on $U$.
        
        As discussed in \autoref{sec:base}, the encoder component of the model is an LSTM that operates at the character-level. We pseudo-train this LSTM with a language model objective on each text sequence $\boldsymbol{x} \in U$.
        
        That is, at each timestep $t$, the LSTM is trained to predict the next character in the input sequence. Given a sequence of characters $\boldsymbol{x}=[x_1, \ldots, x_N]$, the training objective maximizes $\prod_{t=1}^N P(x_t \mid x_{1}, \ldots, x_{t-1})$.
        
        This is the standard LM objective function and has been proven helpful for pretraining LSTMs to improve hidden representations~\cite{10.5555/2969442.2969583,ramachandran-etal-2017-unsupervised}.
        
        \item Train the decoder LSTM with the LM objective described above, using the baseline model's predictions $\{f_\theta({\boldsymbol{x}}) \mid \boldsymbol{x} \in U\}$.
        
        \item Train the sequence-to-sequence model on the \textit{pseudo-annotated dataset} $S$ with the post-correction loss function $\mathcal{L}$ from \autoref{sec:base}.
    \end{enumerate}
    
    \item Given the pseudo-trained model $f_\theta$, \textit{fine-tune} the model on the gold-transcribed dataset $D$, with the loss function $\mathcal{L}$.
    
    \item Repeat step 1 to step 4 until a specified stopping criterion (e.g., no improvement in the validation set performance or reaching the maximum permitted iterations).
\end{enumerate}

As indicated above, self-training is a straightforward semi-supervised technique to leverage documents without gold transcriptions to improve OCR post-correction performance. We note that some self-training methods \cite[\textit{inter alia}]{yarowsky-1995-unsupervised,lee2013pseudo,zoph2020rethinking} replace steps 4 and 5 with a single step that trains $f_\theta$ on $S \cup D$. However, this led to slightly worse performance in our preliminary experiments. We also observed that pseudo-training the LSTMs with an LM objective (steps 3(a) and 3(b) above) is necessary for good performance and that applying the self-training steps on $f_\theta$ from the previous iteration led to better results than re-initializing the model.\footnote{In preliminary experiments, we also tried using $S \cup D$ in step 3(c). However, the post-correction performance was approximately the same as using only the set $S$.}

Further, as recommended in \citet{he2019revisiting} to improve self-training for neural sequence generation, we add a \textit{dropout} layer into the base model at the encoder and decoder hidden states during pseudo-training and fine-tuning (steps 3 and 4).

\section{Lexically-Aware Decoding}
\label{sec:lexical}
Although self-training is a simple approach that leads to improvements in post-correction performance without additional manual annotation, incorrect predictions in the pseudo-annotated data may introduce noise into the model, potentially reinforcing the errors in the next self-training iteration~\cite{zhu2009introduction}. Such noise is more likely to occur in the endangered languages setting, where the base model is trained on minimal data and thus sometimes generates erroneous predictions. 

While some self-training methods use confidence scores to remove noisy predictions (such as \citet{yarowsky-1995-unsupervised}), these are typically designed for classification tasks. Designing such heuristics is challenging for OCR post-correction because the predictions are generated at the character-level; specific characters may be incorrect, but discarding the entire predicted sequence (i.e., the line or sentence) is inefficient, particularly in a low-resource scenario. To mitigate these issues, we propose \textit{lexically-aware decoding}, an inference strategy based on our observations of the challenges associated with the OCR post-correction task. 

More specifically, our preliminary experiments with self-training indicated that the errors made by the model are \textbf{typically inconsistent}. For a particular word, some instances may be correctly predicted by the model. For the instances of the word that are incorrect, we observe that they are likely to be erroneous in different ways, i.e., different subsets of characters in the word are incorrectly predicted. This is expected since the same word can appear in varied contexts, or the first pass OCR for the word can differ. 
Our empirical observations on the pseudo-annotated dataset $S$ showed that, since the errors are inconsistent, \textbf{the correct form of the word is more frequent than incorrect forms}. \textit{Lexically-aware decoding} is designed to influence the OCR post-correction model to generate words that frequently occur in the set $S$, in the expectation that these are correct word forms.

We first describe the construction of a model that accounts for word frequency in the predictions along with a character $n$-gram model to enable the prediction of unseen words. Then, we present a joint decoding method that uses the frequency-based models in combination with the LSTM decoder for improved OCR post-correction. 

\subsection{Count-Based Language Model}
\label{sec:wordlm}

\begin{figure*}
    \centering
    \begin{tabular}{@{}cc@{}}
    \small{(a) Original WFSA} & \small{(b) Minimized WFSA for Known Words} \\
    \resizebox{0.52\textwidth}{!}{%
\begin{tikzpicture}[shorten >=1pt,node distance=2cm,on grid,auto] 
\tikzset{every node/.style={font=\small}}
   \node[state,accepting] (q_0)   {\tiny \texttt{start}}; 
  \node[state] (d2) [right= of q_0]  {}; 
  \node[state] (d1) [below={1.75cm} of d2]  {}; 
  \node[state] (unk) [above={1.75cm} of q_0]  {\small \texttt{unk}}; 
  \node[state] (o1) [right={1.5cm} of d1]  {}; 
  \node[state] (o2) [right={1.5cm} of d2]  {}; 
  \node[state] (g) [right={1.5cm} of o2]  {1}; 
  \node[state] (o22) [right={1.5cm} of o1]  {}; 
  \node[state] (r) [right={1.5cm} of o22]  {2}; 

    \draw[->] (q_0) edge[sloped,pos=0.5, swap]  node {d / 1.6} (d1);
    \draw[->] (q_0) edge[swap]  node {d / 0.3} (d2);
    \draw[->] (q_0) edge[bend left]  node {$\epsilon$ / 3.0} (unk);
    \draw[->] (d1) edge[swap]  node {o} (o1);
    \draw[->] (d2) edge[swap]  node {o} (o2);
    \draw[->] (o1) edge[swap]  node {o} (o22);
    \draw[->] (o2)  edge[swap]  node {g} (g);
    \draw[->] (o22) edge[swap]  node {r} (r);
    \draw[->] (unk) edge[bend left]  node {$\mathcal{B}$} (q_0);
    \draw[->] (g)  edge[bend right, swap]  node {$\mathcal{B}$} (q_0);
    \draw[->] (r)  edge[out=210,in=270]  node {$\mathcal{B}$} (q_0);
\end{tikzpicture}
} & \resizebox{.4\textwidth}{!}{%
\begin{tikzpicture}[shorten >=1pt,node distance=2cm,on grid,auto] 
\tikzset{every node/.style={font=\small}}
   \node[state,accepting] (q_0)   {\tiny \texttt{start}}; 
   \node[state] (d) [right= of q_0]  {1}; 
   \node[state] (o1) [right={1.65cm} of d]  {2}; 
   \node[state] (o2) [above=of g]  {3}; 
   \node[state] (r) [right={1.5cm} of o1]  {4}; 
   
    \draw[->] (q_0) edge  node {d / 0.3} (d);

    \draw[->] (d) edge  node {o} (o1);
    \draw[->] (o1) edge[sloped] node {o / 1.3} (o2);
    \draw[->] (o1)  edge  node {g} (r);
    \draw[->] (o2) edge  node {r} (r);
    \draw[->] (r)  edge[bend left]  node {$\mathcal{B}$} (q_0);
    \node[] () [above={2cm} of d] {\small $\mathcal{B}$: word boundary symbols};
\end{tikzpicture}
} \\
    \end{tabular}
    \caption{The (a) WFSA and (b) minimized WFSA we construct, for a hypothetical language model with a two word vocabulary: $P(\text{dog})=0.75$; $P(\text{door})=0.2$; $P(\text{<unk>})=0.05$. The transition weights are negative log probabilities. In (b), for simplicity, we show only the known word states after determinization and minimization.}
    \label{fig:wfsa}
\end{figure*}
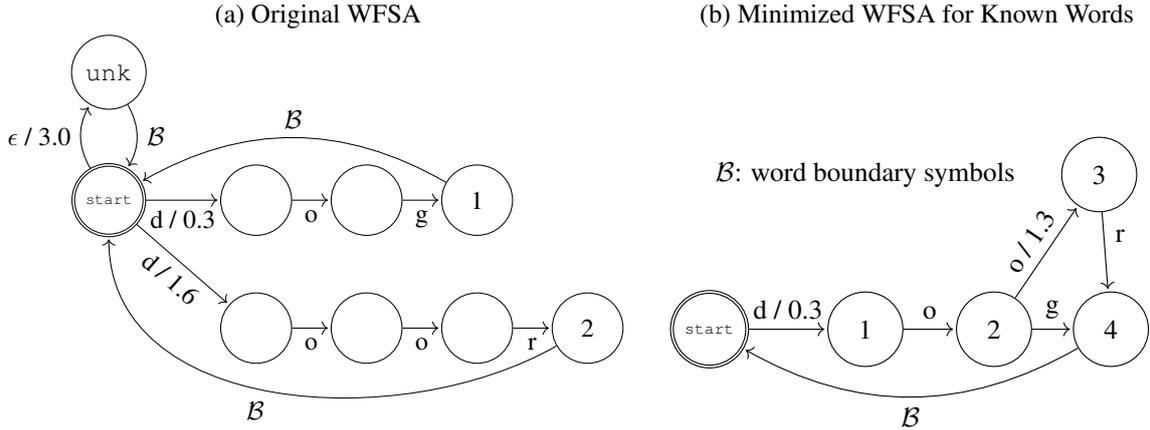

From the self-training method in \autoref{sec:selftraining}, we have a pseudo-annotated dataset $S = \{\langle\boldsymbol{x}, f_\theta(\boldsymbol{x})\rangle \mid \boldsymbol{x} \in U\}$, where $f_\theta(\boldsymbol{x})$ is the model's prediction for input sequence $\boldsymbol{x}$. We train a \textbf{count-based word-level unigram language model} (LM) on $\{f_\theta({\boldsymbol{x}}) \mid \boldsymbol{x} \in U\}$. The LM is built by computing frequency-based probabilities for each word found in the predictions. 

We also have to reserve some probability mass in the LM to account for unknown words (words unseen in the predictions). We use modified Kneser-Ney smoothing to derive the unknown word (\ba\ba <unk>") probability. Since the model is a unigram LM, the smoothing process is similar to absolute discounting. However, we use the discount values based on the modified Kneser-Ney method, which are derived from word counts in the dataset, as opposed to using a fixed discount value~\cite{kneser1995improved, chen1999empirical}. We denote the probability from the smoothed LM for a known word $w$ as $p_\mathrm{word}(w)$ and the unknown word probability as $p_\mathrm{word}(\text{<unk>})$.

A count-based unigram LM is a simple model but is suitable given our empirical observations on word-level errors (described earlier in this section) because (1) it explicitly models word frequency, (2) it is straightforward to update as the pseudo annotated dataset improves over self-training iterations, and (3) it can be expressed as a weighted finite-state automaton which, as we discuss next, has several properties useful for our decoding method.

\subsection{Weighted Finite State Automaton}
\label{sec:wfsa}
A weighted finite-state automaton (WFSA) is a set of states and transitions between the states. Each transition accepts a particular symbol as input and has a weight associated with it. The symbols come from a finite alphabet $\Sigma$. A sequence of consecutive transitions is referred to as a \ba\ba path", and the label of a path is the concatenation of all symbols consumed by its constituent transitions. The WFSA has a start state and a set of final states. A successful path is a path from the start state to a final state, and a sequence of symbols is \ba\ba accepted" by the WFSA if there exists a successful path that consumes this sequence~\cite{mohri2002weighted}.

Since we are focused on decoding and only need the best scoring path for any given sequence (i.e., Viterbi search), we consider the weights over the \textit{tropical semiring}. That is, the weight of a path is the sum of its transition weights, and the score of a sequence of symbols is the minimum weight of all the successful paths that accept that sequence.

Decoding with the post-correction model is at the character-level (\autoref{eq:decoder}), so in order to leverage word frequency in the decoding process, we convert the count-based word-level LM described in \autoref{sec:wordlm} to a WFSA representation that consumes and scores sequences at the character-level.

The WFSA is constructed to accept the \textbf{words known to the LM} by consuming each character in the word (in sequence) as input. The score of the path that accepts a known word $w$ is the negative log of its probability from the LM: $-\log p_\mathrm{word}(w)$. A simple example is shown in \autoref{fig:wfsa}(a).

The WFSA, as described above, can only accept a single word. However, the input and corresponding predictions of the post-correction model are sequences of words, typically lines or sentences. To enable the WFSA to accept such sequences, we add transitions that accept a set $\mathcal{B}$ of word boundary symbols (whitespace, punctuation, and end-of-sequence) from the states at the end of the known words (e.g., states 1 and 2 in \autoref{fig:wfsa}(a)) back to the start state. Once in the start state, the model can begin consuming characters from the next word.

Further, we modify the WFSA such that the start state is also the only final (accepting) state since the predicted sequence is considered complete only when the model predicts an end-of-sequence symbol after the last character.

\paragraph{Character LM for Unknown Words}
To enable the prediction of words unknown to the count-based LM, we include an unknown word state in the WFSA as shown in \autoref{fig:wfsa}(a). We add an $\epsilon$-transition (a transition that consumes no input), with an associated cost $-\log p_\mathrm{word}(\text{<unk>})$ (i.e., the probability mass reserved for unknown words in \autoref{sec:wordlm}) to enter the unknown word state from the start state. The model remains in the unknown state, accepting symbols that form an unseen word, until a word boundary symbol from the set $\mathcal{B}$ is consumed to return to the start state. 

We design the unknown word state to accept any combination of the symbols in $\Sigma$, thereby permitting the prediction of words unseen by the word-level LM. To score each character consumed at the unknown word state, we use a \textbf{character-level $n$-gram language model}.\footnote{We use $n=6$ in this paper. We experimented with different values of $n$ in early experiments but found that $n=6$ gave the best results for all languages in the dataset.} We denote the probabilities from this character $n$-gram LM as $p_\mathrm{char}$. The probability distribution is estimated with modified Kneser-Ney smoothing on character $n$-grams from unique word forms in the set $\{f_\theta({\boldsymbol{x}}) \mid \boldsymbol{x} \in U\}$. We use unique word forms because unknown words are likely rare, and using count-based word forms would undesirably shift the probability mass towards more frequent words.


 
\subsection{Efficient scoring with the WFSA}
\label{sec:scoring}

The constructed WFSA has states to score character sequences that form known words and an unknown word state that relies on a character $n$-gram LM to score unknown sequences.

During inference, we \textbf{independently} score the next character through the known word model and the unknown word model and then choose the best scoring path. This formulation has two advantages: (1) separate scoring allows us to compactly represent the WFSA states for known words and (2) instead of representing the character $n$-gram LM directly in the WFSA, leading to the number of states exponentially increasing with $n$, we can use highly-optimized LM toolkits such as KenLM \cite{heafield-etal-2013-scalable} for scoring unknown words. 

\paragraph{Known Word Model}
Consider the WFSA with only known word states. We apply standard algorithms for determinization and minimization on these states, which leads to an efficient and compact representation of the count-based language model~\cite{10.1017/S135132499600126X}. As shown in \autoref{fig:wfsa}(b), the resultant minimized WFSA has several properties useful for our decoding method, discussed below.

\textbf{Determinization} ensures that each state has at most one outgoing transition that consumes a given input symbol, and \textbf{minimization} eliminates redundant states and transitions, reducing the time and space needed to process an input sequence. 

Further, minimization includes pushing the transition weights towards the start state of the WFSA as much as possible~\cite{mohri2002weighted}. This lends itself well to our method since inference in the OCR post-correction model is performed with \textit{beam search}; if the cost of a path is established closer to the start state, unfavorable hypotheses can be pruned at an earlier timestep, which allows us to avoid errors more effectively within an approximate search algorithm like beam search.

Lastly, since each state in the WFSA has at most one outgoing transition for each symbol, the transition scores can be precomputed and stored as a matrix, allowing efficient retrieval during decoding.

At decoding timestep $t$, let the previous timestep score from the known word model be $\text{known}(y_{t-1})$ and the current WFSA state be $s_{t-1}$. The score for predicting the next character $y_t$ is the weight of the transition from state $s_{t-1}$ that consumes $y_t$ in the minimized WFSA (see \autoref{fig:wfsa}(b)). Thus,
$$\text{known}(y_t) = \text{known}(y_{t-1}) + \text{score}_\mathrm{wfsa}(y_t \mid s_{t-1})$$
where $\text{known}(y_0)=0$. If $y_t$ does not continue the path of any known word, then $\text{score}_\mathrm{wfsa}(y_t)$ is $\inf$.

\paragraph{Unknown Word Model}
We use the probability $p_\mathrm{char}$ from the character $n$-gram language model to score unknown words. In general, at decoding timestep $t$, the unknown model score for $y_t$ will be: 
{\small 
\begin{equation*}
\text{unk}(y_t)= \text{unk}(y_{t-1})-\log p_\mathrm{char}(y_t \mid y_{t-n}, \dots, y_{t-1})
\end{equation*}}
\normalsize
\noindent
However, if $y_{t-1} \in \mathcal{B}$ (i.e., the previous word is complete) or $t=0$, the WFSA is currently in the start state. To begin an unknown word, we also need to add the weight of entering the unknown word state to $\text{unk}(y_t)$, i.e., $-\log p_\mathrm{word}(\text{<unk>})$.

\paragraph{Best Scoring Path}
The scores are in the tropical semiring (negative log probabilities). At timestep $t$, the best score for $y_t$ from the lexical models is:
\begin{equation}
\text{score}_\mathrm{lex}(y_t) = \min(\text{known}(y_t),\text{unk}(y_t))
\label{eq:lex}
\end{equation}
\noindent

During decoding, we keep track of both the known and unknown model scores for the \textit{current word} being generated in the hypothesis. When the word is completed (when $y_t \in \mathcal{B}$), both the known and unknown word models return to the start state of the WFSA (see \autoref{fig:wfsa}). Since the two paths are in the same state and are thus indistinguishable with respect to future predictions in the hypothesis, we choose the best scoring path to continue decoding. This is known as hypothesis recombination.


The WFSA framework, thus, allows us to efficiently represent the word-level LM in a manner that scores symbols at the character-level and leverage a character $n$-gram model to score unknown words. This enables joint inference with the character-level LSTM decoder in the OCR post-correction model, as discussed below.

\subsection{Joint Decoding with the LSTM}
\label{sec:jointdecoding}

At decoding timestep $t$, let $p_\mathrm{lstm}(y_t)$ be the probability of generating a character $y_t$ based on the LSTM decoder's hidden state (\autoref{eq:decoder}). We also compute $\text{score}_\mathrm{lex}(y_t)$, which is a negative log probability, as defined in \autoref{eq:lex}. The final probability of predicting $y_t$ is obtained through linear interpolation between these two scores,\footnote{We leave other interpolation techniques like log-linear interpolation and more complex combinations of scores (such as the WFSA-based reranking and rescoring methods proposed by \citet{ryskina-etal-2021-comparative}) as potential future work.} weighted by a hyperparameter $\lambda$:
\begin{equation}
p(y_t) = (1-\lambda) \cdot p_\mathrm{lstm}(y_t) + \lambda \cdot p_\mathrm{lex}(y_t)
\label{eq:joint}
\end{equation}
\noindent
where $p_\mathrm{lex}(y_t) = \exp\left(-\text{score}_\mathrm{lex}(y_t)\right)$.

This joint decoding strategy is applied when performing inference with beam search using a trained OCR post-correction model. When used in combination with self-training, the predictions made by the model improve as we repeat the self-training process, iteratively improving the count-based LM and resulting in a better distribution of $p_\mathrm{lex}(y_t)$.

\section{Experiments}
\label{sec:expts}

In this section, we present experiments with our semi-supervised post-correction method on four typologically diverse endangered languages. 

\subsection{Datasets}
\label{sec:dataset}

We use the OCR post-correction dataset from \citet{rijhwani-etal-2020-ocr} which contains transcribed documents in three endangered languages: Ainu, Griko, and Yakkha. Additionally, in this paper, we create a similar dataset in the endangered language Kwak'wala. We describe the datasets below, including the sizes of the gold transcribed and unannotated sets we use for semi-supervised training:

\textbf{Ainu} (\texttt{ain}) is a severely endangered language from northern Japan. The dataset contains pages from a book of Ainu epic poetry~\cite{kindaichi1931ainu}. The Ainu text is written in the Latin script. The dataset contains 816 manually transcribed lines as well as 7,646 lines without gold transcriptions.

\textbf{Griko} (\texttt{grk}), an endangered Greek dialect spoken in southern Italy, is written with a combination of Latin and Greek alphabet. The document in the dataset is a book of Griko folk tales~\cite{stomeo1980racconti}. There are 807 and 3,084 sentences with and without gold transcriptions, respectively.

\textbf{Yakkha} (\texttt{ybh}) is an endangered language spoken in Nepal and is written in the Devanagari script. The dataset contains transcriptions of three children's books~\cite{yakkha-elar}. In total, there are 159 manually transcribed sentences and no unannotated lines in the dataset. Therefore, as the unannotated set, we use the first pass OCR on the validation and test sets in a transductive learning setting ($\approx30$ sentences: see \autoref{sec:setup} for data splits).

\textbf{Kwak'wala} (\texttt{kwk}) is spoken on Northern Vancouver Island, nearby small islands, and the opposing mainland. The language is severely endangered, with estimates of $\approx$150 first-language speakers, all over the age of 70. The Kwak'wala language includes 42 consonantal phonemes (twice as many as English) and a wide range of allophonic vowels. Several writing systems exist and community preference varies between two orthographies: the U’mista and Liq’wala systems. 

However, much of the written documentation for Kwak’wala is in another orthography that was developed by anthropologist Franz Boas. The Boas orthography~\cite{boas1900sketch} was used in the extensive documentation of the Kwak'wala language and its speakers produced by Boas in collaboration with native-speaker George Hunt. The Boas writing system uses Latin script characters as well as diacritics and digraphs to represent phonemic differences. Although the Boas orthography is not widely used today, the cultural and linguistic materials previously written by Boas are of tremendous value to community-based researchers. However, they are minimally accessible since they currently exist only as non-searchable scanned images.

In consultation with members of language revitalization projects in three Kwakiutl communities (Tsulquate, Fort Rupert, Quatsino), we focus on digitizing these significant cultural resources. We create a dataset with pages from the \ba\ba Ethnology of the Kwakiutl''~\cite{boas1921ethnology}, containing 262 gold-transcribed lines and 2,255 unannotated lines.

\subsection{Experimental Setup}
\label{sec:setup}

\paragraph{Data Splits} We follow \citet{rijhwani-etal-2020-ocr} and perform 10-fold cross-validation for all experiments. For each language, the gold-transcribed data is split into 10 segments, and for each cross-validation fold, eight segments are used for training, one for validation, and one for testing.

\paragraph{Metrics} We evaluate our systems in terms of character error rate (CER) and word error rate (WER), both standard metrics for measuring OCR and OCR post-correction performance~\cite{berg-kirkpatrick-etal-2013-unsupervised,schulz-kuhn-2017-multi}. CER is the character-level edit distance between the predicted text and the corresponding gold transcription, divided by the total number of characters in the gold transcription. WER is similar but is calculated at the word-level. For readability, we report CER and WER as percentages for all experiments.

\paragraph{Methods} 
In our experiments, we compare the performance of the following methods:

\begin{itemize}[leftmargin=*,itemsep=0pt]
    \item \textsc{First-Pass}: To obtain the first pass OCR transcription, we experiment with two existing OCR systems: Google Vision~\cite{fujii2017sequence} and Ocular~\cite{berg-kirkpatrick-etal-2013-unsupervised}.
    
    For each language, we choose the best performing first pass system, the details of which are in \autoref{sec:firstpass}. We use Ocular for Kwak'wala and Google Vision for Ainu, Griko, and Yakkha.
    
    \item \textsc{Base}: The current state-of-the-art in OCR post-correction for endangered language texts (\citet{rijhwani-etal-2020-ocr}; described in \autoref{sec:base}).
    \item \textsc{Semi-Supervised}: Our proposed method as described in \autoref{sec:selftraining} and \autoref{sec:lexical}.
\end{itemize}

\paragraph{Implementation}
The neural post-correction models are implemented using the DyNet neural network toolkit~\cite{dynet}. The WFSA is implemented using the MFST Python wrapper on OpenFST~\cite{francislandau2020mfst}, and we use the KenLM toolkit~\cite{heafield-etal-2013-scalable} to train and query the character $n$-gram language model. Following \citet{rijhwani-etal-2020-ocr}, results reported are the average of five randomly seeded runs (i.e., five runs for each of the 10 cross-validation folds).

\begin{table*}[tb]
    \centering
    \small
    \begin{tabular}{@{}l|cccc|cccc@{}}
    \toprule
        & \multicolumn{4}{c|}{\% Character Error Rate} &\multicolumn{4}{c}{\% Word Error Rate} \\[0.3em]
        Model & \multicolumn{1}{c}{\texttt{ain}} & \multicolumn{1}{c}{\texttt{grk}} & \multicolumn{1}{c}{\texttt{ybh}} & \multicolumn{1}{c|}{\texttt{kwk}} & \multicolumn{1}{c}{\texttt{ain}} & \multicolumn{1}{c}{\texttt{grk}} & \multicolumn{1}{c}{\texttt{ybh}} & \multicolumn{1}{c}{\texttt{kwk}} \\
        \midrule
        \textsc{First-Pass} & $1.34$ & $3.27$ & $8.90$ & $7.90$ & $6.27$ & $15.63$ & $31.64$ & $38.22$ \\
        \textsc{Base} & $0.80$ & $1.70$ & $8.44$ & $4.97$ & $5.19$ & $7.51$ & $21.33$ & $27.65$ \\
        \midrule
        \textsc{Semi-Supervised} & & & & & & \\
        \ \ Self-Training & $0.82$ & $1.45$ & $7.20$ & $4.00$ & $5.31$ & $6.47$ & $18.09$ & $23.98$ \\
        \ \ Lexical Decoding & $0.81$ & $1.51$ & $7.56$ & $4.28$ & $5.18$ & $6.60$ & $19.13$ & $25.09$ \\
        \ \ Both & $\boldsymbol{0.63}$ & $\boldsymbol{1.37}$ & $\boldsymbol{5.98}$ & $\boldsymbol{3.82}$ & $\boldsymbol{4.43}$ & $\boldsymbol{6.36}$ & $\boldsymbol{16.65}$ & $\boldsymbol{22.61}$ \\
        \midrule
        Error Reduction $\left(\frac{\textsc{Base}-\text{Both}}{\textsc{Base}}\right)$ & 21\% & 19\% & 29\% & 23\% & 15\% & 15\% & 22\% & 18\% \\
    \bottomrule
    \end{tabular}
    \ifspace
    \vspace{-1em}
    \fi
    \caption{Our semi-supervised approach improves performance over the baselines (10-fold cross-validation averaged over five randomly seeded runs). \ba\ba Self-Training'' and \ba\ba Lexical Decoding'' refer to experiments where we use these methods independently. \ba\ba Both'' refers to their combination. We \textbf{highlight} the best model for each language.}
    \label{tab:results}
    \ifspace
    \vspace{-1em}
    \fi
\end{table*}

\begin{table*}[tb]
    \centering
    \small
    \begin{tabular}{@{}l|cccc|cccc@{}}
    \toprule
        & \multicolumn{4}{c|}{\% Character Error Rate} &\multicolumn{4}{c}{\% Word Error Rate} \\[0.3em]
        OCR System & \multicolumn{1}{c}{\texttt{ain}} & \multicolumn{1}{c}{\texttt{grk}} & \multicolumn{1}{c}{\texttt{ybh}} & \multicolumn{1}{c|}{\texttt{kwk}} & \multicolumn{1}{c}{\texttt{ain}} & \multicolumn{1}{c}{\texttt{grk}} & \multicolumn{1}{c}{\texttt{ybh}} & \multicolumn{1}{c}{\texttt{kwk}} \\
        \midrule
        Ocular & $10.49$ & $4.58$ & $75.60$ & $\boldsymbol{7.90}$ & $47.47$ & $15.71$ & $99.37$ & $\boldsymbol{38.22}$ \\
        Google Vision & $\boldsymbol{1.34}$ & $\boldsymbol{3.27}$ & $\boldsymbol{8.90}$ & $21.12$ & $\boldsymbol{6.27}$ & $\boldsymbol{15.63}$ & $\boldsymbol{31.64}$ & $82.08$ \\
    \bottomrule
    \end{tabular}
    \ifspace
    \vspace{-1em}
    \fi
    \caption{First pass OCR system performance. If the language's script is not covered by Google Vision (as for Kwak'wala), then Ocular results in better recognition. Otherwise, Google Vision OCR is usually significantly better.}
    \label{tab:fpresults}
    \ifspace
    \vspace{-1em}
    \fi
\end{table*}


\subsection{Main Results}

\autoref{tab:results} shows the performance of the baselines and our proposed semi-supervised approaches for the four languages in the dataset. For all languages, using semi-supervised learning leads to substantial reductions in both CER and WER.

We note that we did a hyperparameter search over the number of self-training iterations and the weight of the WFSA $\lambda$, and \autoref{tab:results} presents the best models based on the validation set WER. Extensive analysis of these factors is in \autoref{subsec:analysis}.

First, we note that the \textsc{Base} post-correction method improves error rates over the first pass for all languages. With our proposed semi-supervised learning method, combining self-training with lexically-aware decoding leads to the best performance across all the languages, with error rate reductions in the range of 15\%-29\%.

This is especially noticeable in Ainu, where using either self-training or lexical decoding independently results in worse performance than the \textsc{Base} system, but jointly using them improves the CER by 21\%. For the other languages, the independent components improve over the base model but less so than their combination. This indicates the complementary nature of the two components: the language model used for lexically-aware decoding is improved by self-training. In turn, it reinforces correctly predicted words to counteract the influence of incorrect pseudo-annotated instances.

\begin{table*}[tb]
    \centering
    \small
    \begin{tabular}{@{}ll|cccc|cccc@{}}
    \toprule
        Known Word& Unknown Word & \multicolumn{4}{c|}{\% Character Error Rate} &\multicolumn{4}{c}{\% Word Error Rate} \\
        Model & Model & \multicolumn{1}{c}{\texttt{ain}} & \multicolumn{1}{c}{\texttt{grk}} & \multicolumn{1}{c}{\texttt{ybh}} & \multicolumn{1}{c|}{\texttt{kwk}} & \multicolumn{1}{c}{\texttt{ain}} & \multicolumn{1}{c}{\texttt{grk}} & \multicolumn{1}{c}{\texttt{ybh}} & \multicolumn{1}{c}{\texttt{kwk}} \\
        \midrule
        \textsc{CharLM} & \textit{(not needed)} & $0.64$ & $1.43$ & $6.22$ & $3.85$ & $4.50$ & $6.44$ & $16.78$ & $22.90$ \\
        \textsc{WordLM} & Character uniform & $0.64$ & $1.42$ & $6.12$ & $3.95$ & $4.50$ & $6.39$ & $16.71$ & $23.11$ \\
        \textsc{Ours} & Character $n$-gram &  $\boldsymbol{0.63}$ & $\boldsymbol{1.37}$ & $\boldsymbol{5.98}$ & $\boldsymbol{3.82}$ & $\boldsymbol{4.43}$ & $\boldsymbol{6.36}$ & $\boldsymbol{16.65}$ & $\boldsymbol{22.61}$ \\
    \bottomrule
    \end{tabular}
    \ifspace
    \vspace{-1em}
    \fi
    \caption{A more informed unknown word model (character $n$-gram) in combination with the word-level known word model consistently performs better than the alternatives for all four languages in our dataset.}
    \label{tab:lmresults}
    \ifspace
    \vspace{-1em}
    \fi
\end{table*}

\subsection{First Pass OCR Systems}
\label{sec:firstpass}

We experiment with two existing OCR systems to obtain a first pass transcription on our dataset. The first of these is the Google Vision system~\cite{fujii2017sequence,ingle2019scalable}. This off-the-shelf OCR model supports 60 languages in 27 scripts -- these are primarily higher-resourced languages and do not include our target endangered languages.

The second system is Ocular~\cite{berg-kirkpatrick-etal-2013-unsupervised}. Ocular uses a generative model to transcribe scanned documents: the model generates the image by learning the font of the document. Ocular relies on a character n-gram language model trained on the target language. We initialize the LM with the small number of gold-transcribed pages in our dataset. For this, we use the 10-fold cross-validation setup described in \autoref{sec:setup}: we use the training segments to train the Ocular LM and the test segment to evaluate OCR performance. The font model has parameters to learn the shape of each character in the LM vocabulary. After initialization, the parameters are updated in an unsupervised manner with EM until convergence.

Results are presented in \autoref{tab:fpresults}. We note that although the Google OCR system is not trained on our target languages, it is trained on large amounts of data in high-resource languages that share writing systems with Ainu, Griko, and Yakkha (Latin, Greek, Devanagari scripts) and thus, can recognize characters in these scripts with reasonable accuracy (see \citet{rijhwani-etal-2020-ocr} for a more detailed analysis). On the other hand, the performance is much worse on Kwak'wala since the system has not been trained on the Boas orthography, which is unique to the Kwak'wala language. The Boas orthography uses several Latin script characters, which the system is able to recognize, but it also includes characters unique to the writing system that are incorrectly transcribed by the OCR model.

We find that the performance on Kwak'wala is considerably better with the Ocular system because the LM is trained on Kwak'wala text. Thus, unlike Google Vision, the model vocabulary contains the Boas writing system's alphabet. On the other hand, Ocular's performance on Ainu and Griko is worse than Google Vision, likely due to the minimal data available for training it. Moreover, we observe that performance is correlated with the \textit{word overlap} between test data and the data used for training the LM, demonstrating Ocular's reliance on a strong language model -- the word overlap is 73\% for Griko, 56\% for Kwak'wala, and 48\% for Ainu.

Finally, we find that Ocular does not perform well on the Yakkha dataset. This is because the design of Ocular's font model does not work with how the Devanagari script is written. More specifically, when a vowel diacritic is applied to a consonant, the characters are combined: e.g., {\raisebox{-1.5pt}{\includegraphics[height=9.5pt]{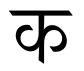}}} + {\raisebox{-1.5pt}{\includegraphics[height=9.5pt]{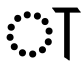}}} = {\raisebox{-1.5pt}{\includegraphics[height=9.5pt]{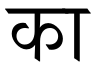}}}. In Unicode, this is represented by two characters \ba\ba {\raisebox{-1.5pt}{\includegraphics[height=9.5pt]{deva_images/k.pdf}}}" and \ba\ba {\raisebox{-1.5pt}{\includegraphics[height=9.5pt]{deva_images/a.pdf}}}'', where the dotted circle is the \textit{character combination marker} in the Unicode Standard.\footnote{\url{https://www.unicode.org/versions/Unicode13.0.0/ch02.pdf}}

However, since Ocular's font model operates at the character-level, it tries to generate the images of these two characters separately. Generating the diacritic \ba\ba {\raisebox{-1.5pt}{\includegraphics[height=9.5pt]{deva_images/a.pdf}}}'' on its own is not meaningful: the dotted circle never appears in the input image because it is supposed to be combined. Thus, the font model is unable to converge as it cannot handle character combinations when generating the image.

\subsection{Comparing Language Models}

Our proposed decoding method uses a count-based word-level LM in combination with a character $n$-gram LM to compute $p_\mathrm{lex}$ for joint decoding with the LSTM decoder (\autoref{eq:joint}). In this section, we substitute this model with two other variants of count-based LMs to compute $p_\mathrm{lex}$ and evaluate their performance:
\begin{itemize}[leftmargin=*]
    \item \textsc{CharLM}: We use a character 6-gram language model trained on the model predictions from self-training $\{f_\theta({\boldsymbol{x}}) \mid \boldsymbol{x} \in U\}$, estimated with modified Kneser-Ney smoothing.
    \item \textsc{WordLM}: We use the word-level LM described in \autoref{sec:wordlm}, but do not use a character $n$-gram model for unknown words. Instead, we score unknown words with a simple uniform probability over all characters in the vocabulary.
\end{itemize}

We tune $\lambda$ on the validation set for each model independently and report results with the best setting in \autoref{tab:lmresults}. Using either \textsc{CharLM} or \textsc{WordLM} for lexically-aware decoding improves the error rates with respect to the \textsc{Base} model. The word-level model performs better for all languages except Kwak'wala, likely due to the large percentage of unknown words in this language. We also see that our proposed method, which leverages a count-based word-level LM for known words combined with a character-level LM for scoring unknown words, results in the best performance overall.

Although not observed in our dataset, we note that some printed materials have a high degree of spelling variation or contain texts for which word tokenization is difficult. In such cases, the word-level model may not be as effective, but \textsc{CharLM} can still be used with the proposed lexically-aware decoding framework to obtain improved performance over the baseline method.

\subsection{Analysis}
\label{subsec:analysis}
\renewcommand{\arraystretch}{1.2}
\begin{table}[tb]
    \centering
    \small
    \begin{tabular}{@{}lc|c@{ }c|c@{ }c@{}}
    \toprule
        Lang.&LM& \multicolumn{2}{c|}{Known} & \multicolumn{2}{c}{Unknown}\\
        Code & Coverage & \textsc{Base} & \textsc{Ours} & \textsc{Base} & \textsc{Ours} \\
        \midrule
        \texttt{ain} & $0.97$ & $0.95$ & $0.98$ & $0.08$ & $0.25$ \\
        \texttt{grk} & $0.94$ & $0.89$ & $0.96$ & $0.51$ & $0.71$ \\
        \texttt{ybh} & $0.68$ & $0.90$ & $0.95$ & $0.51$ & $0.59$ \\
        \texttt{kwk} & $0.59$ & $0.89$ & $0.92$ & $0.50$ & $0.58$ \\
        \midrule
        Average & $0.80$ & $0.91$ & $\boldsymbol{0.95}$ & $0.40$ & $\boldsymbol{0.53}$ \\
    \bottomrule
    \end{tabular}
    \caption{Our method improves over the base model on words that are both known and unknown to the WFSA. We show the fraction of known test words, and the fraction of correctly predicted known and unknown words.}
    \label{tab:words}
    \ifspace
    \vspace{-1em}
    \fi
\end{table}
\renewcommand{\arraystretch}{1.0}

\pgfplotstableread[row sep=\\,col sep=&]{
grikoX & grikoCER & grikoWER \\
0 & 1.45 & 6.47 \\
0.025 & 1.62 & 6.52 \\
0.05 & 1.37 & 6.36 \\
0.1 & 2.61 & 11.79 \\
0.25 & 3.72 & 17.73 \\
0.5 & 8.27 & 27.06 \\
}\grcdata

\pgfplotstableread[row sep=\\,col sep=&]{
kwakX & kwakCER & kwakWER \\
0.1 & 4.69 & 26.38 \\
0.075 & 4.7 & 26.55 \\
0.05 & 4.72 & 26.66 \\
0.0375 & 4.72 & 26.71 \\
0.025 & 4.73 & 26.77 \\
0.0125 & 4.74 & 26.87 \\
}\kwadata

\pgfplotstableread[row sep=\\,col sep=&]{
yakkhaX & yakkhaCER & yakkhaWER \\
0.025 & 6.29 & 17.03\\
0.05 & 5.98	& 16.65\\
0.1 & 6.26 & 16.78 \\
0.25 & 6.69 & 19.75\\
0.5 &  10.49 & 22.71\\
}\yakdata


\begin{figure}[t]
\centering
\begin{tabular}{c}
\begin{tikzpicture}[trim left=0.2\textwidth,trim right=0cm]
    \begin{axis}[
    title={Griko},
    every axis plot post/.style={/pgf/number format/fixed},
    width=0.48\textwidth,
            height=3cm,
            ymajorgrids=false,
            yminorgrids=false,
            xmode=log,
            log ticks with fixed point,
            every x tick label/.append style={font=\tiny},
            every y tick label/.append style={font=\tiny},
            xtick={025,0.05,0.1,0.5},
            ymin=0,
            tick pos=left,
            axis y line*=left,
            axis x line*=bottom,
            ylabel near ticks,
            ylabel shift={-3pt},
            xlabel near ticks,
            xlabel shift={-3pt},
            enlarge x limits=0.05,
            title style={yshift=-.5cm,font=\footnotesize},
    ylabel={\scriptsize \% WER},xlabel={\scriptsize WFSA Weight $\lambda$ (log scale)}]
    \addplot [mark=none,nodes near coords,
            every node near coord/.append style={font=\tiny,color=black},] table[x=grikoX,y=grikoWER]{\grcdata};
    \addplot [only marks,mark=square*,mark options={fill=white,draw=white},mark size=1pt,] table[x=grikoX,y=grikoWER]{\grcdata};
    \addplot [only marks,mark=o,mark options={fill=black,draw=black},mark size=.3pt,] table[x=grikoX,y=grikoWER]{\grcdata};
    \end{axis}
\end{tikzpicture}\\[0.5em]
\begin{tikzpicture}[trim left=0.2\textwidth,trim right=0cm]
    \begin{axis}[
    title={Yakkha},
    every axis plot post/.style={/pgf/number format/fixed},
    width=0.48\textwidth,
            height=3cm,
            ymajorgrids=false,
            yminorgrids=false,
            xmode=log,
            log ticks with fixed point,
            every x tick label/.append style={font=\tiny},
            every y tick label/.append style={font=\tiny},
            xtick={025,0.05,0.1,0.5},
            ymin=0,
            tick pos=left,
            axis y line*=left,
            axis x line*=bottom,
            ylabel near ticks,
            ylabel shift={-3pt},
            xlabel near ticks,
            xlabel shift={-3pt},
            enlarge x limits=0.05,
            title style={yshift=-.5cm,font=\footnotesize},
    ylabel={\scriptsize \% WER},xlabel={\scriptsize WFSA Weight $\lambda$ (log scale)}]
    \addplot [mark=none,color=black,nodes near coords,every node near coord/.append style={font=\tiny,yshift=-10pt}] table[x=yakkhaX,y=yakkhaWER]{\yakdata};
    \addplot [only marks,mark=square*,mark options={fill=white,draw=white},mark size=1pt,] table[x=yakkhaX,y=yakkhaWER]{\yakdata};
    \addplot [only marks,mark=o,mark options={fill=black,draw=black},mark size=.3pt,] table[x=yakkhaX,y=yakkhaWER]{\yakdata};
    \end{axis}
\end{tikzpicture}
\end{tabular}
    \caption{The weight of the WFST during joint decoding can affect word error rate (sometimes significantly, as in Griko; top). All other hyperparameters are kept equal and correspond to the best systems in each language.}
    \label{fig:fstfactor}
    \ifspace
    \vspace{-1em}
    \fi

\end{figure}
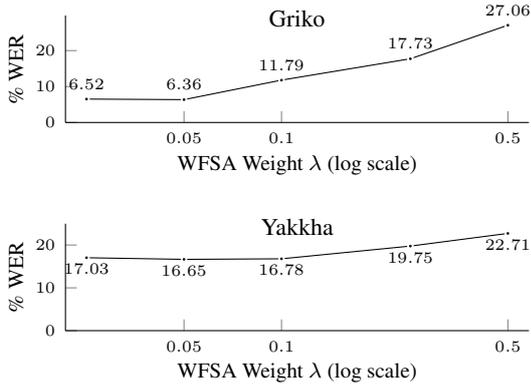
\pgfplotstableread[row sep=\\,col sep=&]{
Iterations & AinuNoLex & GrikoNoLex & YakkhaNoLex & KwakNoLex & AinuLex & GrikoLex & YakkhaLex & KwakLex \\ 
0&  5.19 & 7.51 & 21.3 & 32.3 & 5.18 & 7.47 & 20.2 & 29.7\\
1& 5.31 & 7.34 & 18.5 & 29.1 & 4.65 & 7.26 & 17.2 & 27.7\\	
2& 5.88 & 6.56 & 18.4 & 28.0 & 4.73 & 6.51 & 17.0 & 27.1\\
3& 5.62 & 6.83 & 18.8 & 27.8 & 4.81 & 6.88 & 16.6 & 26.7\\
4& 6.05 & 6.67 & 18.1 & 27.4 & 4.43 & 6.36 & 17.4 & 26.7\\
5& 6.07 & 6.47 & 19.1 & 27.5 & 4.50 & 6.53 & 18.2 & 26.4\\
}\itdata

\begin{figure}[t]
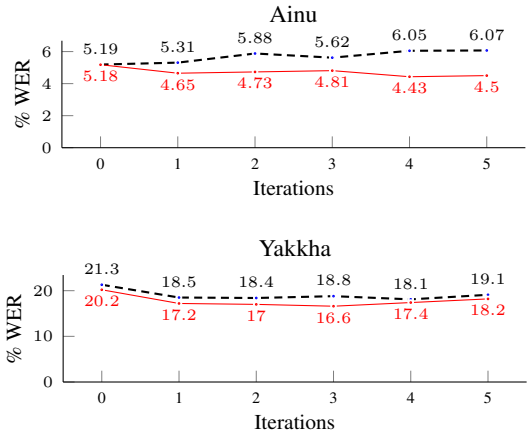

\centering
\begin{tabular}{@{}c@{}}
\niceplot{AinuNoLex}{AinuLex}{\footnotesize Ainu}{\scriptsize{\% WER}}{\scriptsize{Iterations}} \\[.5em]
\niceplot{YakkhaNoLex}{YakkhaLex}{\footnotesize Yakkha}{\scriptsize{\% WER}}{\scriptsize{Iterations}}
\end{tabular}
    \ifspace
    \vspace{-1.5em}
    \fi
    \caption{Integrating lexically-aware decoding through interpolation with a WFSA (\textcolor{red}{red lines}) aids self-training in improving WER across iterations. Black dashed lines correspond to self-training without lexical decoding.}
    \label{fig:iterations}
    \ifspace
    \vspace{-1em}
    \fi
\end{figure}


We analyze specific components of our model to understand the advantages of our proposed approach.

\paragraph{Known vs. Unknown Words}
We first identify the source of the improvements that our approach makes over the baseline. \autoref{tab:words} presents the fraction of correctly predicted words, split on whether these words are ``known" to the WFSA (i.e., in the vocabulary of the word-level LM) or ``unknown". Intuitively, we expect that decoding with the WFSA will improve prediction on the known words.

Compared to the baseline, our method improves on words known to the WFSA, moving from 91\% to 95\% accuracy on average. Our method also improves unknown word prediction over the baseline from an average accuracy of 40\% to 53\%. In cases like Kwak'wala, where, due to the rich morphology of the language, more than 40\% of the test words are unseen, including an unknown word model in the WFSA is particularly important.

\paragraph{WFSA Weight}
One of the important hyperparameters of our lexically-aware method is the weight that we place on the WFSA score during inference ($\lambda$ in \autoref{eq:joint}). Specifically, in the case of Griko, we find that the value of this hyperparameter can significantly affect performance. As shown in \autoref{fig:fstfactor}, high weights of $\lambda$ (i.e., more weight on the WFSA) lead to suboptimal WER, while lower $\lambda$ leads to much better performance.

This hyperparameter is less important in the other three languages, leading to smaller variations in performance. As an example, we depict the effect on Yakkha in \autoref{fig:fstfactor}, where increasing $\lambda$ does not affect performance as much as in Griko.

\paragraph{Self-Training Iterations} 
The evolution of WER across~5 self-training iterations for Ainu and Yakkha is shown in \autoref{fig:iterations}. Particularly for Ainu, we see that combination with lexically-aware decoding is crucial for the success of self-training. For Yakkha, self-training does improve performance independently but is more effective when lexically-aware decoding is used (error rates on Griko and Kwak'wala follow a similar trend).

\pgfplotstableread[row sep=\\,col sep=&]{
uX & uCER & uWER \\
0.125 & 0 & 7.42 \\
0.25 & 0 & 6.82 \\
0.5 & 0 & 6.48 \\
1.0 & 1.37 & 6.36 \\
}\unlabdata

\pgfplotstableread[row sep=\\,col sep=&]{
lX & lCER & lWER \\
0.125 & 0 & 12.02 \\
0.25 & 0 & 10.25 \\
0.5 & 0 & 7.10 \\
1.0 & 1.37 & 6.36 \\
}\labdata

\pgfplotstableread[row sep=\\,col sep=&]{
lX & lCER & lWER \\
0.125 & 0 & 15.98 \\
0.25 & 0 & 13.02 \\
0.5 & 0 & 8.03 \\
1.0 &  & 7.51 \\
}\labdatabase

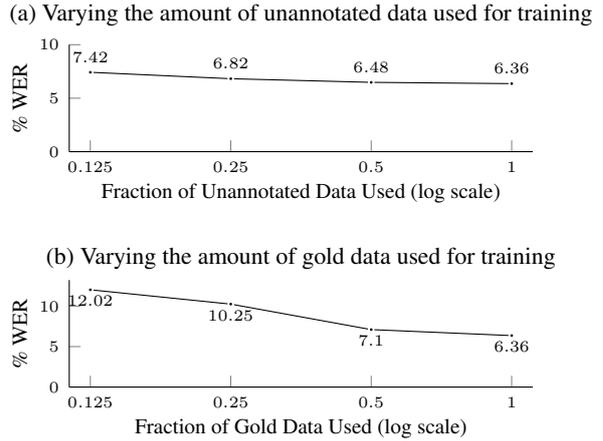
\begin{figure}[t]
\centering
\begin{tabular}{c}
\begin{tikzpicture}[trim left=0.2\textwidth,trim right=0cm]
    \begin{axis}[
    title={(a) Varying the amount of unannotated data used for training},
    every axis plot post/.style={/pgf/number format/fixed},
    width=0.48\textwidth,
            height=3cm,
            ymajorgrids=false,
            yminorgrids=false,
            xmode=log,
            log ticks with fixed point,
            every x tick label/.append style={font=\tiny},
            every y tick label/.append style={font=\tiny},
            xtick={.125,.25,.5,1.0},
            ytick={0,5,10},
            ymax=10,
            ymin=0,
            tick pos=left,
            axis y line*=left,
            axis x line*=bottom,
            ylabel near ticks,
            ylabel shift={-3pt},
            xlabel near ticks,
            xlabel shift={-3pt},
            enlarge x limits=0.05,
            title style={yshift=-.1cm,font=\footnotesize},
    ylabel={\scriptsize \% WER},xlabel={\scriptsize Fraction of Unannotated Data Used (log scale)}]
    \addplot [mark=none,nodes near coords,
            every node near coord/.append style={font=\tiny,color=black},] table[x=uX,y=uWER]{\unlabdata};
    \addplot [only marks,mark=square*,mark options={fill=white,draw=white},mark size=1pt,] table[x=uX,y=uWER]{\unlabdata};
    \addplot [only marks,mark=o,mark options={fill=black,draw=black},mark size=.3pt,] table[x=uX,y=uWER]{\unlabdata};
    \end{axis}
\end{tikzpicture}\\[.5em]
\begin{tikzpicture}[trim left=0.2\textwidth,trim right=0cm]
    \begin{axis}[
    title={(b) Varying the amount of gold data used for training},
    every axis plot post/.style={/pgf/number format/fixed},
    width=0.48\textwidth,
            height=3cm,
            ymajorgrids=false,
            yminorgrids=false,
            xmode=log,
            log ticks with fixed point,
            every x tick label/.append style={font=\tiny},
            every y tick label/.append style={font=\tiny},
            xtick={.125,.25,.5,1.0},
            ytick={0,5,10},
            ymin=0,
            tick pos=left,
            axis y line*=left,
            axis x line*=bottom,
            ylabel near ticks,
            ylabel shift={-3pt},
            xlabel near ticks,
            xlabel shift={-3pt},
            enlarge x limits=0.05,
            title style={yshift=-.2cm,font=\footnotesize},
    ylabel={\scriptsize \% WER},xlabel={\scriptsize Fraction of Gold Data Used (log scale)}]
    \addplot [mark=none,color=black,nodes near coords,every node near coord/.append style={font=\tiny,yshift=-10pt}] table[x=lX,y=lWER]{\labdata};
    \addplot [only marks,mark=square*,mark options={fill=white,draw=white},mark size=1pt,] table[x=lX,y=lWER]{\labdata};
    \addplot [only marks,mark=o,mark options={fill=black,draw=black},mark size=.3pt,] table[x=lX,y=lWER]{\labdata};
    \end{axis}
\end{tikzpicture}

\end{tabular}
    \caption{Even a small amount of unannotated data is useful for our semi-supervised method, improving WER over \textsc{Base} (WER=7.51) in (a). Varying the size of gold-annotated data has a stronger effect on post-correction performance in (b). Results are shown with Griko.}
    \label{fig:datasize}
    \ifspace
    \vspace{-1em}
    \fi

\end{figure}

\paragraph{Dataset Size}
We study the effect of varying the amount of gold-transcribed and unannotated data used for training. The WER when varying the size of the Griko datasets is shown in \autoref{fig:datasize} (the size of each set is varied while keeping the other set at its full size). We see that reducing the amount of gold-transcribed data worsens WER significantly. On the other hand, reducing the unannotated data has a smaller effect: even with a little unannotated data, our method improves over the \textsc{Base} model.

\begin{figure*}[tb]
    \centering
    \small
    \begin{tabular}{@{}lcc@{\qquad \qquad}cc@{}}
    & \multicolumn{2}{l}{\qquad \quad Errors \textit{fixed} by our method} & \multicolumn{2}{c}{Errors \textit{introduced} by our method}\\[.05cm]
        & (a) Griko & (b) Kwak'wala & (c) Yakkha & (d) Kwak'wala \\[.05cm]
        \raisebox{0.35em}{[Image]} & \frame{\includegraphics[height=1.2em]{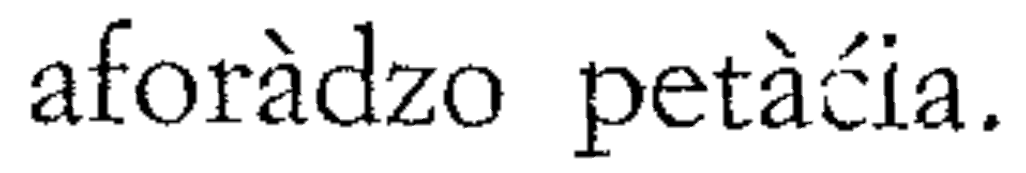}} & \includegraphics[height=1.1em]{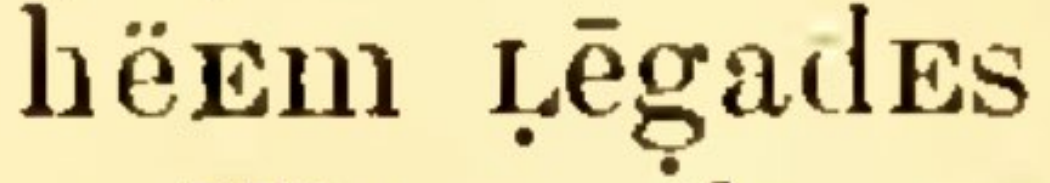} & \frame{\includegraphics[height=1.3em]{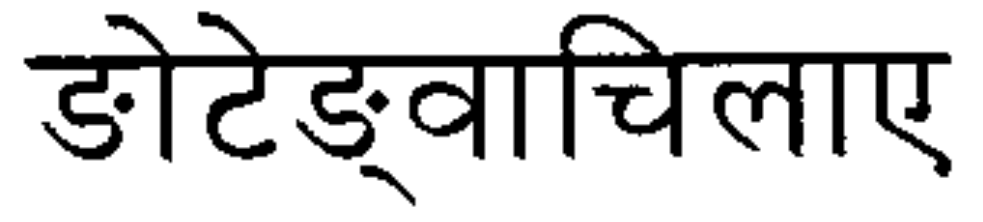}} & \includegraphics[height=1.2em]{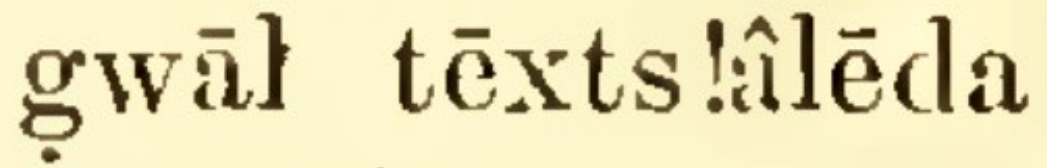}  \\
        
        & \multicolumn{2}{l}{\hspace{1.1cm} $\boldsymbol{\downarrow}$ \hspace{2.5cm} $\boldsymbol{\downarrow}$} & \multicolumn{2}{l}{\hspace{0.6cm} $\boldsymbol{\downarrow}$ \hspace{2.3cm} $\boldsymbol{\downarrow}$}\\
        
        \raisebox{0.3em}{[First pass OCR]} & \raisebox{0.25em}{\normalsize aforàdzo petà\textcolor{burntred}{\textbf{c}}ia.} &
        \includegraphics[height=1em]{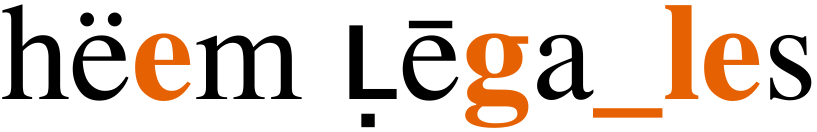} & \includegraphics[height=1.15em]{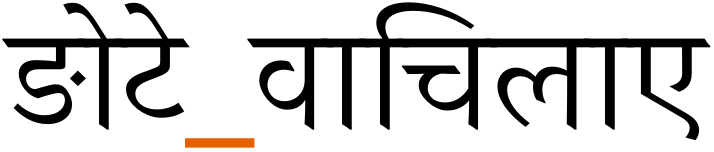} & \includegraphics[height=1em]{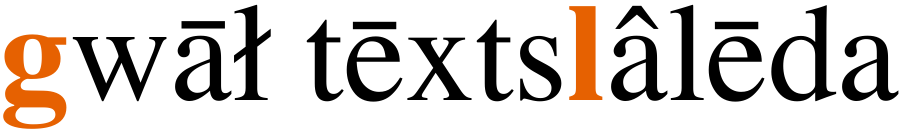}\\
        
        & \multicolumn{2}{l}{\hspace{1.1cm} $\boldsymbol{\downarrow}$ \hspace{2.5cm} $\boldsymbol{\downarrow}$} & \multicolumn{2}{l}{\hspace{0.6cm} $\boldsymbol{\downarrow}$ \hspace{2.3cm} $\boldsymbol{\downarrow}$}\\
        
        \raisebox{0.5em}{[Post-corrected \textsc{Base}]} & \raisebox{0.5em}{\normalsize aforà\textcolor{burntred}{\textbf{\d{d}}}zo petà\textcolor{burntred}{\textbf{c}}ia.} &
        \raisebox{0.25em}{\includegraphics[height=1em]{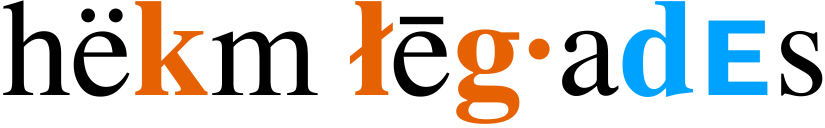}} &  \raisebox{0.1em}{\includegraphics[height=1.35em]{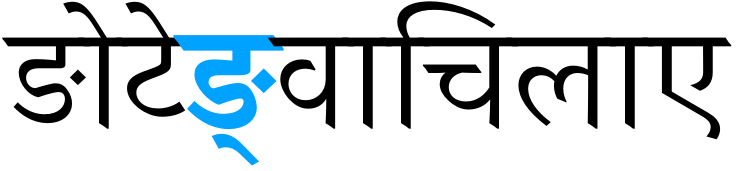}} & \raisebox{0.1em}{\includegraphics[height=1.25em]{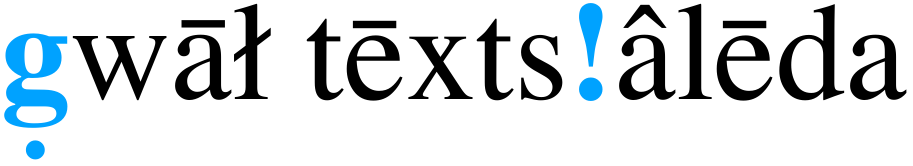}}\\[.05cm]
        
        \raisebox{0.5em}{[Post-corrected \textsc{Ours}]} & \raisebox{0.5em}{\normalsize aforàdzo petà\textcolor{burntblue}{\textbf{ć}}ia.} & 
        \includegraphics[height=1.35em]{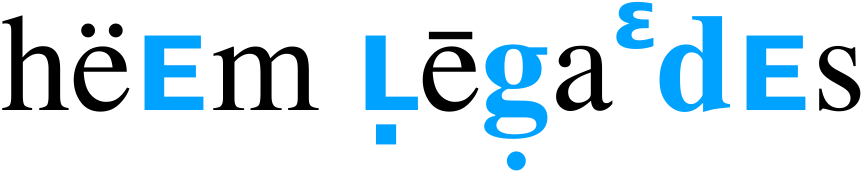} &  \raisebox{0.1em}{\includegraphics[height=1.35em]{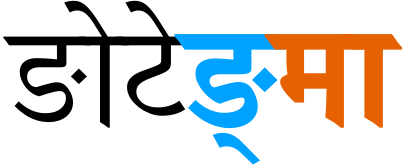}} & \raisebox{0.1em}{\includegraphics[height=1.25em]{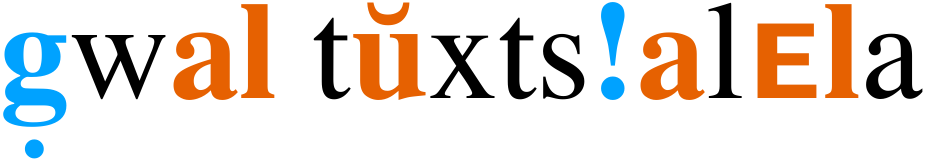}}
    \end{tabular}
    \caption{Our post-correction model \textcolor{burntblue}{\textbf{fixes}} many of the first pass OCR \textcolor{burntred}{\textbf{errors}} that the base model does not fix such as (a) and (b). In rare cases, our method introduces \textcolor{burntred}{\textbf{errors}} into the transcription such as (c) and (d).}
    \label{fig:qualitative}
\end{figure*}

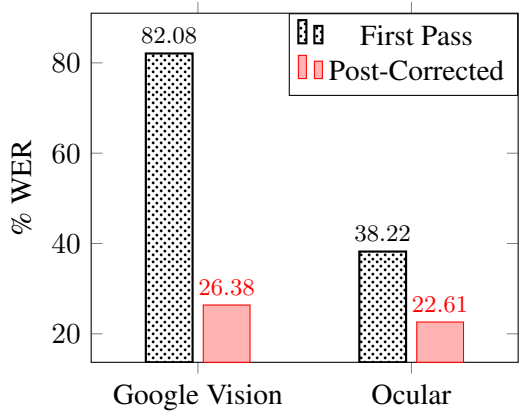
\begin{figure}[tb]
\centering
\begin{tikzpicture}[trim left=-0.25cm]
\begin{axis}[
    width=0.45\textwidth,
    ybar=4pt,
    enlargelimits=0.15,
    legend style={at={(1,1)},
      anchor=north east},
    ylabel={\% WER},
    symbolic x coords={Google Vision, Ocular},
    xtick=data,
    nodes near coords,
    nodes near coords align={vertical},
    ylabel near ticks,
    bar width=17.5pt,
    enlarge x limits=0.5,
    every node near coord/.append style={font=\footnotesize},
    ]
\addplot
[thick,pattern=crosshatch dots, pattern color=black]
coordinates {(Google Vision,82.08) (Ocular,38.22)};
\addplot coordinates {(Google Vision,26.38) (Ocular,22.61)};
\legend{First Pass, Post-Corrected}
\end{axis}
\end{tikzpicture}
\caption{Our post-correction model significantly improves recognition accuracy over different first pass OCR systems that have varied error rates (Google Vision and Ocular). Results are shown with Kwak'wala.}
\label{fig:fpcorr}
\end{figure}

\paragraph{Error Rate in the First Pass OCR} To evaluate how the error rate in the first pass OCR transcription affects subsequent post-correction, we measure the performance of our proposed method when applied to first pass outputs from two OCR systems: Google Vision and Ocular (described in \autoref{sec:firstpass}). \autoref{fig:fpcorr} shows the WER on the Kwak'wala dataset. We see that, although Google Vision has a much higher first pass error rate than Ocular, the post-correction model improves performance over both OCR systems. We also note that the relative error reduction is higher for the Google Vision system (68\%) than for Ocular (41\%), likely because the Ocular LM is trained on the same data as the post-correction model.

\paragraph{Qualitative Analysis}
In \autoref{fig:qualitative}, we show examples of errors fixed as well as errors introduced by our post-correction model, as compared to the baseline system. In \autoref{fig:qualitative} (a) and (b), we see that although the baseline corrects some of the errors in the first pass OCR, it also introduces errors such as extra diacritics and incorrect substitutions. Using our proposed method leads to an error-free transcription of these images. However, in \autoref{fig:qualitative} (c) and (d), we see that our method occasionally introduces errors in predictions. Specifically, although the model fixes the first pass errors, it generates words that are considerably different from the target. Such errors likely occur when the model follows an incorrect path in the WFSA during lexically-aware decoding. Since we are using beam search, the correct path cannot be recovered if it was pruned at an earlier timestep.

\section{Related Work}
OCR post-correction is well-studied in the high-resource setting, particularly for English. Recent methods primarily use neural encoder-decoder models~\cite{dong-smith-2018-multi,icdar,hamalainen-hengchen-2019-paft}. There has been relatively little work on lower-resourced languages. \citet{kolak-resnik-2005-ocr} present a probabilistic edit distance model for post-correction on Cebuano and Igbo, and \citet{krishna-etal-2018-upcycle} use a sequence-to-sequence model with a copy mechanism for improved performance on Romanized Sanskrit OCR.

While existing neural post-correction methods do not rely on lexical information, some earlier methods use dictionaries to improve performance. For example, \citet{tong-evans-1996-statistical} and \citet{niklas2010unsupervised} use lexicons in combination with $n$-gram context to generate post-correction candidates for erroneous words. These methods are typically evaluated on English and assume the presence of high-coverage lexicons~\cite{schulz-kuhn-2017-multi}, making them difficult to adapt to endangered languages.

Related to our decoding method are models that incorporate lexical knowledge into neural machine translation. \citet{arthur-etal-2016-incorporating} propose adding a dictionary for translating low-frequency words and \citet{zhang-etal-2018-guiding} improve decoding by upweighting translations that contain relevant words. Additionally, there are methods which add \textit{hard} lexical constraints by forcing predictions to contain user-specified words and phrases~\cite{hokamp-liu-2017-lexically,post-vilar-2018-fast}.

Lastly, we note that our proposed approach combines information from a neural model and a finite-state machine to leverage the advantages of both. In a similar direction, \citet{rastogi-etal-2016-weighting} and \citet{lin-etal-2019-neural} design finite state architectures with paths weighted by contextual features from an LSTM. These methods use joint parameterizations of the models and are thus more complex to train (particularly in the low-resource setting) than the joint decoding method we propose in this paper.

\section{Conclusion}

Digitization at scale for documents in under-represented languages is a promising avenue towards tackling one aspect of their marginalization, the lack of data. With this work, we take a step towards better digitization for extremely data-scarce scenarios. We develop a semi-supervised method that combines self-training with lexically-aware decoding, reducing error rates by up to 29\% over a state-of-the-art OCR post-correction model on four typologically diverse endangered languages.

In future work, we plan to expand our method to take advantage of additional outputs of the language documentation process. For example, documentary linguists typically collect word lists (which range from lists of common words like the Swadesh lists~\cite{swadesh1955towards} to domain-specific vocabularies). Using such word lists within the lexically-aware decoding framework could further improve performance and enable the application of our technique to even lower-resourced languages.

Additionally, the improvements we achieve through semi-supervised learning are potentially orthogonal to the improvements \citet{rijhwani-etal-2020-ocr} achieve by incorporating information from translations of the target text.
As future work, we plan to investigate the combination of these two approaches in an attempt to utilize all available sources of information to improve performance. 


Finally, while using a character-level n-gram LM improves performance on unknown words, it does not explicitly utilize morphological structure to generate unseen inflections of words. In the future, we plan to incorporate morphological analysis during post-correction decoding, which will be helpful for morphologically rich endangered languages.


\section*{Acknowledgments}

Shruti Rijhwani was supported by the Bloomberg Data Science Ph.D. Fellowship for this work.

This work was also supported by grant PR-276810-21 (\textit{``Unlocking Endangered Language Resources"}) from the National Endowment for the Humanities, grant 1761548 (\textit{``Discovering and Demonstrating Linguistic Features for Language Documentation''}) from the National Science Foundation, National Research Council Indigenous Language Technology grant \textit{\ba\ba Kwak’wala Corpus Collection Project''}, and Government of Canada Social Sciences and Humanities Research Council Insight Development grant GR002807 \textit{\ba\ba \b K'\b an\b k'otła\b x\b ants \b Awi'na\b gwis (Knowing our land)''}. Any views, findings, conclusions or recommendations expressed in this publication do not necessarily represent those of the National Endowment for the Humanities. 

We would like to thank Jaymyn La Vallee for assisting with the Kwak'wala dataset annotation as well as Samridhi Choudhary, Siddharth Dalmia, Deepak Gopinath, Maria Ryskina, the reviewers, and the Action Editors for feedback on the paper.

\bibliography{tacl2018}
\bibliographystyle{acl_natbib}

\end{document}